# The acquisition of English irregular inflections by Yemeni L1 Arabic learners: A Universal Grammar approach

Muneef Y. Alsawsh, Department of English & Translation, Ibb University
alshawushmuneef@gmail.com
Mohammed Q. Shormani, Department of English & Translation, Ibb University
shormani@ibbuniv.edu.ye/ https://orcid.org/0000-0002-0138-4793

**Abstract**

This study examines the acquisition of English irregular inflections by Yemeni learners of English as a second language (L2), utilizing a Universal Grammar (UG) approach. Within the UG approach, the study considers Feature Reassembly Hypothesis (FRH) (Lardiere, 2008, 2009) part of UG, focusing on the roles of first language (L1) transfer and L2 developmental influence. It analyzes learner errors across two developmental stages. Stage 1 data reveal a dominant influence of L1 transfer, particularly in phonological and structural mismatches, while stage 2 data demonstrate increased learner sensitivity to UG properties and morphological reconfiguration toward the target language. Findings reveal that errors in irregular inflectional morphology are attributed to both interlingual and intralingual sources, with overgeneralization of L2 rules as a common developmental strategy. Statistical analysis, including a one-way ANOVA, indicates significant improvement in the production of well-formed irregular inflections from stage 1 to stage 2, underscoring learners' continued access to UG. However, persistent difficulties with consonant change, zero-morpheme, and -a plural inflections suggest that limited exposure, ineffective input modeling, and insufficient instructional quality constrain full UG access. The study concludes that while L1 transfer and L2 developmental factors influence initial stages of acquisition, appropriate linguistic input and instruction are critical for facilitating UG-driven feature reassembly in adult L2 learners.

**Keywords**: L2 acquisition, irregular inflections, Universal Grammar, L1 transfer, Feature Reassembly

## 1. Introduction

Acquiring a second language has long been a central and often contentious topic within various linguistic frameworks (Alshawsh & Shormani, 2024). This paper addresses this phenomenon within Universal Grammar (UG) approach, considering Feature Reassembly Framework (Lardiere, 2008, 2009) part of UG. The concept of UG has been one of the ideas in Chomsky's generative framework. The generative framework could be said to have passed through three major stages, constituting three broader frameworks, viz., Phrase Structure Rules (PSRs), Principles and Parameters (P&P) framework and minimalism (see e.g., Shormani, 2024). PSRs could be traced back to the introduction of Chomsky's *Syntactic Structures* (1957). It was the first attempt to define the structure of natural language through explicit rules and a generative mechanism. This model marks the beginning of UG as a formal system designed to generate all and only the grammatical sentences of a language. In this framework, UG was conceived as a finite set of PSRs and transformational rules that account for the syntax of any natural language (see e.g., Chomsky, 1957; Ouhalla, 1999; Radford, 2009; Shormani, 2013, 2024a). These rules generate deep structures, which are then transformed to surface structures through syntactic operations. Such rules are language-specific, and UG has been seen as a blueprint that could be parameterized differently for each language. However, this early formulation of UG was criticized for being too complex, language-particular, and lacking a clear account of how children acquire language so rapidly and uniformly (Shormani, 2014, et seq, 2015).

The second framework in which UG develops is P&P Framework. The P&P model, first developed in Chomsky's *Lectures on Government and Binding* (1981) and subsequent research in the early 1990s, marked a major evolution in Chomsky's linguistic theory. Focusing on sets of specific rules for each language, this framework proposes that UG consists of universal principles shared by all languages and a finite set of parameters whose values vary across languages (see e.g., Chomsky, 1981, 1982; Shormani, 2014*et seq*). In this framework, UG was seen as a biologically endowed cognitive faculty consisting of: i) Principles as invariant properties of syntax (e.g., structure dependence, locality conditions, binding theory), and ii) Parameters as binary switches that account for cross-linguistic variation (e.g., pro-drop vs. non-pro-drop languages, head-initial vs. head-final structures). The third framework, Minimalism, seeks to reduce linguistic theory to the most economical and computationally efficient principles consistent with UG (see e.g., Chomsky, 2000, et seq; Shormani, 2014, et seq). Minimalism aims to derive the richness of linguistic phenomena from a small set of primitive operations, primarily Merge, and general cognitive principles. UG in minimalism is seen as a highly economical, minimal system containing only what is necessary for language. There are four main mental operations constituting UG in minimalism: Select, Merge, Agree, and Spell-out.

In P&P, children are thought to be born with access to UG, and language acquisition is conceptualized as a process of parameter setting based on the linguistic input they receive. Once a parameter is set, it constrains future syntactic derivations. UG is universal and innate, explaining the uniformity of L1 acquisition. Parameter setting accounts for cross-linguistic diversity with a limited cognitive system (see e.g., Ouhalla, 1999; Shormani, 2024). UG in P&P made language acquisition more plausible by drastically reducing the learning burden on the child.

In this study, we aim to study the acquisition of irregular inflections by Yemeni learners of English. We also aim to examine whether these learners still have access to UG in acquiring English irregular inflections. The study thus is concerned with identifying the barriers and factors constraining such access to UG. To address the gap of this study, we examine the acquisition of English irregular inflectional morphemes: consonant change, vowel +consonant change, zero-morpheme, -a (plural), suppletion, -en (plural) and vowel change. All these inflections have been examined via asking the participants to respond to the written test consisting of four parts: error identification and correction test, plural forming test, free composition test to write a complete sentence expressing past event/action test and free composition test in a form of writing a paragraph about any topic of the five given topics (see Appendix II).

There are few studies examining the role of UG in acquiring L2 inflections cross-linguistically. In the Arab/Yemeni context, there is a study conducted on the acquisition of L2 inflections by (Alshawsh & Shormani, 2024) in which they tried to identify the UG role in the this process, to the best of our knowledge, no study has been conducted on the acquisition of English irregular inflections by Yemeni or Arab learners. In this study, 30 Yemeni Arabic-speaking learners of English were randomly chosen as the participants of this study. Data were collected in two stages, the first while they were at level 3, and the second while they were at level 4 of their university study. Thus, this study strives to answer the following questions:

1. To what extent do Yemeni learners of English have access to UG while acquiring English irregular inflections?
2. What are the problems encountered by Yemeni learners in acquiring English irregular inflections?
3. What are the possible solutions to overcome, or at least, mitigate these problems.

## 2. Theoretical foundations

### 2.1. UG and Inflections

Given the three frameworks of UG development introduced above, a question arises as whether UG encodes inflections and vice versa in language acquisition process. To answer this question, it is worth mentioning that inflectional morphology provides direct evidence of the grammatical features that UG makes available to all human languages. When a child or second language learner acquires inflections, they are not just memorizing word forms; they are mapping abstract grammatical features (e.g., [+past], [+plural]) onto specific morphophonological forms (like "-ed" in English or "-s"). This mapping reflects UG's provision of universal grammatical categories and constraints. In L1 acquisition, UG enables learners to set parameters based on linguistic input. For example, English marks tense with inflectional morphemes like "-ed," whereas Chinese does not inflect verbs for tense. The child uses UG to select and fix the parameter: "Does my language use overt tense morphology?" In this way, inflections reflect how parameters are set. However, in L2 acquisition the learner may already have parameters fixed from L1, and acquiring inflections involves resetting these parameters and reassembling the grammatical features based on the L2 system (Lardiere, 2008). For instance, an L1 Arabic learner of English may struggle with subject-verb agreement inflections in English if such agreement is realized differently in Arabic.

In minimalism, UG includes mental operations such as: i) *Select* picks up lexical items from the mental lexicon, which contains words along with their syntactic, semantic, and phonological features. These selected items serve as the raw materials for sentence formation. Once lexical items are selected, ii) the operation *Merge* applies. *Merge* is the core structure-building operation in syntax. It takes two syntactic objects and combines them into a new set, creating hierarchical structure. This process is recursive and enables the construction of complex expressions. There are two types of *Merge*:

*External Merge*, which combines two separate elements (e.g., *the* and *dog* to form *the dog*), and *Internal Merge*, which moves an element already in the structure to a new position, such as in forming questions (e.g., moving *what* to the beginning of *What did you eat?*), iii) the operation *Agree*, may be the operation concerning us more than others in this study, handles the interaction of grammatical features between different elements in a sentence. It operates between a *probe* (an element with unvalued features, like Tense) and a *goal* (an element with matching valued features, such as the subject noun). Through Agree, features such as number, gender, and case are checked and valued. For example, in the sentence *The dogs run*, the Tense node probes for number features and agrees with the plural subject *dogs*, ensuring grammatical agreement, and iv) the operation *Spell-out* is the operation that transfers the syntactic structure to the interfaces for pronunciation and interpretation, specifically, the Phonological Form (PF) and Logical Form (LF).

In minimalism, there is also another part of UG, viz., the Interfaces: internal and external interfaces. UG interfaces with two other systems: the conceptual-intentional system (semantics) and the articulatory-perceptual system (phonology). Features: Lexical items are bundles of formal features (e.g., φ-features, case, tense) that must be checked and valued in the syntax. Rather than parameter setting, the feature assembly favors feature selection and assembly (for L1 acquisition) and feature reassembly (for L2 acquisition). This change reflects a move away from binary parameter values toward a more flexible, feature-driven model (see e.g., Chomsky, 2005; Lardiere, 2008, 2009; Shormani, 2015). Within the Feature Reassembly Hypothesis (FRH), UG provides a universal inventory of features, but languages differ in how they package and realize these features. Inflectional errors in L2 learners often result from misalignments in how features are bundled. For example, in English, [+plural] is typically marked with "-s," but in other languages, pluralization might also encode gender or case as in the case of Arabic (see e.g., Aoun et al., 2010; Shormani, 2013). Learners must reconfigure their internal feature systems to correctly produce L2 inflections, showing how inflection relates to feature mapping under UG constraints. Thus, in terms of errors in L1 and L2 acquisition, inflectional morphology errors such as overgeneralization (e.g., *goed*, *childs*) or omission (e.g., *He go yesterday*) reflect developmental stages governed by UG. Such errors are not random but systematic, suggesting that learners are testing hypotheses about how UG manifests in the target language (see e.g., Gass & Slinker, 2008; Shormani, 2010, 2012a & b, et seq). These patterns provide empirical evidence that learners are drawing on innate UG-guided mechanisms during acquisition.

### 2.3. English Irregular inflections

Concerning irregular inflections, English utilizes this type of inflections across a variety of word categories such as nouns and verbs, for tense, number and gender, for example, it also employs internal changes to the base form—such as consonant and vowel changes, suppletion, and zero change. In consonant change, spend-spent can be taken as an example, and in the case of verbs, i.e. past tense, and ox-oxen in the case of nouns, i.e. number, viz., plural. In vowel change, speak-spoke and *man → men*, are good examples. Suppletion can be seen in nouns as person-people, and go-went in verbs. Xero morpheme can be found in examples as in 'sheep & deer', ' deer' and 'fish', **cut-cut** in the case of nouns and verbs, respectively (Berent et al., 2002; Shormani, 2013)

There are also cases in which two operations such as Vowel + Consonant Change occur. For example, Vowel + Consonant Change is an irregular inflection the core function of which is to express past feature as in 'speak➜ spoken' & teach➜ taught'. This shows that the morphological realization of these irregular inflection can never be subjected to a rule as the regular counterpart that is rule-based concatenating the stem with the suffix inflection, and "is insensitive to the phonological content of the words they stand for" (Berent et al., 2009, p. 463); instead, such inflections are stored in the lexicon with their idiosyncratic properties though they may manifest regular-based rules as suffixation as it is the case with 'dwarf➜ dwarves'.

**3. Acquisition of English irregular inflections**

Knowing that syntactic and semantic features are universal, functional morphology remains the linguistic domain that L2 learners need to acquire as it the locus of parametric variation (Lardiere, 2008, 2009, Slabokova, 2003, 2008 & 2016; Jensen et al., 2020). Because of this, the acquisition of inflections is assumed to be the most formidable and hardest task learners face as such learning task is hardly master or it may end up with a failure. Resetting parameter in the feature-selection sense, the P&P account, is too narrow to account for L2 acquisition as it never makes clear why is morphological variability the most encountering learning problem L2 learners face. Lardiere (2008, p. 7) argues that such an account "might be necessary for successful acquisition but would clearly not be anywhere near sufficient", and this is clearly realized with the L2 learners' inability to acquire the morphological competence. Based on this, they become aware of "precisely which forms go with' which features" (p. 4) and hence, be familiar with "the ways in which grammatical features are morphologically combined and conditioned may well affect their acquirability and overt realization in SLA" (p. 4). As a result, Lardiere (2008 & 2009) emphasizes that the resetting parameter in the feature-selection sense (see the *representational deficit approach* (see Hawkins 2003, Hawkins & Liszka 2003, Tsimpli 2003; see also Tsimpli & Roussou, 1991; Hawkins & Chan, 1997 for more details) should be enriched with "an adequate description of feature- reassembly in addition to feature-selection" (p. 26).

The FRH is a feature-based approach; it necessitates the mastery of the morphological competence in acquiring English inflectional features. Such competence can make it clear for L2 learners about what constitutes the formal features, how they are constrained and conditioned in their contexts, what constitutes an obligatory context for these features, etc. (Lardiere, 2008, 2009). Empirical evidence of a learning problem is best provided by Lardiere's (2008) + past feature showing how such feature is selected and grammaticalized in English, Irish, and Somali as this feature neither expresses "a unitary interpretable feature" (p. 6) nor be restricted to a specific domain. For instance, the **+past** feature, in English, is not confined to expressing an action/state completed before the moment of speech as in '*Ali went to the university yesterday'*; instead, it can also express the perfective aspect as in '*She met Ali in the street'*, irrealis mood in conditionals as in ' *She wished if she was in his situation*', and so on (see Lardiere, 2008, p. 6, for more details). In Irish, **+past** feature is realized within the CP projection as a complementizer in an agree relation with the embedded clause's past tense whereas in Somali, this feature is located in DPs on determiners and adjectives in nominal DPs signaling not only 'past time' agreement, but it can also express the temporal habitualness, evidentiality, and alienable possession in predicative genitive constructions (see Lardiere, 2008, pp. 6-7, for examples and details). It is obviously clear that assembling the +past feature in the three languages varies in terms of the condition, factors, and contexts on its encoding.

Accordingly, neither the resetting parameter in feature-selection sense nor the representational deficit approach can adequately account for the variation in encoding the +past feature without knowing the obligatory or optional conditions, restrictions, and context among others. Based on the representational deficit account, English L2 learners of Somali do not need to reset this parameter as encoding such feature selection in Somali is realized in two distinct 'selection events', and therefore, such learners "still needed to parametrically select the nominal [past] feature", and thus, they would face a "formidable learning task" in the mastery of "the obligatory or optional conditions and restrictions on its overt expression" (p. 8). Thus, in L1 and L2 acquisition errors in inflectional morphology such as overgeneralization (e.g., *goed*, *childs*) or omission (e.g., *He go yesterday*) reflect developmental stages governed by UG. Such errors are not random but systematic, suggesting that learners are testing hypotheses about how UG manifests in the target language (Shormani, 2014, et seq). These patterns provide empirical evidence that learners are drawing on innate UG-guided mechanisms during acquisition of English inflections.

## 4. Methodology

The nature of this study is to a large extent longitudinal. It is carried out in two stages: the first stage (Henceforth, stage 1) occurs when the participants were at level three of their university education, whereas the second (Henceforth, stage 2) takes place when the same participants were at level four. The stage 1 is primarily concerned with examining L1 transfer role in L2 English inflection acquisition, while the stage 2 is set for examining the UG role, and secondly on comparing each's role to the other.

### 4.1. Participants

This study involves 30 Yemeni Arabic-speaking learners of English, studying at the department of English, Faculty of Arts, Ibb University, Yemen. These 30 randomly selected learners participated twice: first while they were at level 3, i.e. Stage 1 of data collection took place in 2023. When they reached level 4, the same group also participated in Stage 2 of data collection in 2024.

### 4.2. Study design

The study data were collected via objective and subjective tests. The data collection process was carried out in two distinct stages. During the first stage, participants were enrolled in their third academic level, first semester [The 10$^{th}$ of September 2023] while in the second stage, they had progressed to the fourth level, also in their first semester [The 7$^{th}$ of August 2024]. At Stage 1, participants were invited to voluntarily complete a written test composed of four sections, as outlined in Section 4.2. In Stage 2, the same group of participants was asked to respond to the identical written test administered during the first stage. Data from each stage were recorded independently. The participants' responses were examined by the researcher in collaboration with the academic supervisor. Analysis involved basic frequency counts and percentages. The data were categorized into two primary types: well-formed and ill-formed constructions. The ill-formed responses were further classified into four subcategories: omission, addition, misselection, and left undone (cf. e.g. Park et al., 2021; Al-Jadani, 2016; Ntalli, 2021). Responses falling under the "left undone" category were excluded from the final corpus. This study assumes that well-formed inflectional constructions reflect participants' level of access to UG properties. Conversely, ill-formed constructions are interpreted as indicative of either L1 transfer or L2 influence, manifested through learning strategies such as rule overgeneralization or the excessive application of specific inflectional rules to inappropriate contexts (see also Alshawsh & Shormani, 2024).

**4.2.1. Study Instrument**

This study instrument begins by Informed Consent Form for ethical issues (see Appendix I). It consists of a test in its two types: the objective and subjective tests, to collect data from the participants (see Appendix II). The tests were sent to four referees for validation. These referees' specializations include linguistics, literature, and TESOL. Their suggestions were taken into careful consideration and incorporated into the final version of the tests. The subjective tests include Error identification and the plural forming whereas the objective test includes a Free composition test in its two types: viz, sentence building test using 10 verbs to express either an ongoing situation or past action or event and writing a paragraph in not more than 150 words about one of five topics. The Error Identification test focuses on measuring whether Arabic L2 learners of English have the English morphological/inflectional knowledge which make them able to identify the inflectional error or they have not yet acquired it let alone identifying at what stage those learners have arrived in acquiring the English inflectional system. This test is like the suppliance in obligatory context technique utilized by Hwang and Lardiere (2013), and Slabakova (2016) to measure the participants competence to provide the correct inflection in contexts where it is obligatory. Regarding the plural forming test, it aims at measuring the learners' ability to internalize the L2 underlying knowledge of the plural feature. The test relevance lies in providing an analysis of the patterns of language development the participants manifest in the acquisition of English inflections on the one hand and in identify and analyzing plural errors and the underlying source, i.e., L1 transfer or L2 influence on the other hand.

As for the sentence building test, it is intended to measure the participants' acquisition of the underlying knowledge of some verbs so that they express them meaningfully and with the correct form. This test is relevant to the study concerns as it can provide it with empirical data so that it is possible to analyze these data to measure the participants' proficiency in producing any factual or habitual action or event and past action or event using specific items. Writing a paragraph test is focused on identifying the participants' inflectional errors and areas of difficulty and success they face while writing a paragraph. The relevance of this test to the study concern lies in measuring the participants' use of inflection in meaningful context, and this would certainly enable the researcher to decide how much of UG and L1 transfer those participants exhibit in their production.

## 5. Results and Discussion

### 5.1. Results

So far discussed, L2 learners are mostly observed to encounter learning difficulties in the acquisition of L2 inflections as such categories are the locus of parametric variations. These learning difficulties are obviously observed in the errors they produce which in turn signal the failure of L2 learners to remap/reconfigure the already mapped formal features, L1features, and those available from UG into new or different configurations in the L2 (Lardiere, 2008, 2009; Slabakova, 2008, 2009). Put simply, the errors L2 learners make indicate that such learners have not yet acquired the morphological competence or the knowledge of these inflections, i.e., bundles a of semantic, syntactic and phonological features that build up the meaning and acceptability of the whole sentence (Slabakova, 2008). We now turn to the results of the study presented in tables (1&2) collected in two stages according to the purpose, i.e. testing the L1 transfer and L2 interference, and to demonstrate the role of UG properties in acquiring irregular inflections. A One-way ANOVA analysis was used to identify differences between groups and also to identify differences within groups as presented in tables (3). The results are tabulated as follows.

Table (1) presents results of stage 1 in terms of irregular inflection. As can be seen, we have (7) irregular inflections. There are 288 well-formed occurrences of inflections, and 193 ill-formed occurrences of inflections distributed as (8) ill-formed instances of omission, 108 instances of addition, 14 instances of misselection and 63 instances of left undone.

**Table 1: Results collected from the participants at stage 1**

| Items | Irregular Inflection | | | | | | | | | |
|---|---|---|---|---|---|---|---|---|---|---|
| | Well-formed | | Ill-formed | | | | | | | |
| | | | Omission | | Addition | | Misselection | | Left undone | |
| | F | % | F | % | F | % | F | % | F | % |
| Consonant change | 41 | 14.2 | 6 | 75 | 8 | 7.4 | 2 | 14.3 | 3 | 4.8 |
| Vowel +Consonant change | 37 | 12.8 | 0 | 0 | 12 | 11.1 | 2 | 14.3 | 9 | 14.3 |
| Zero-morpheme | 69 | 24 | 1 | 12.5 | 35 | 32.4 | 3 | 21.4 | 12 | 19 |
| -A (plural) | 5 | 1.7 | 0 | 0 | 40 | 37 | 2 | 14.3 | 13 | 20.6 |
| Suppletion | 25 | 8.7 | 0 | 0 | 0 | 0 | 2 | 14.3 | 3 | 4.8 |

| | | | | | | | | | |
|---|---|---|---|---|---|---|---|---|---|
| -en (plural) | 29 | 10.1 | 0 | 0 | 1 | 0.9 | 0 | 0 | 0 | 0 |
| Vowel change | 82 | 28.5 | 1 | 12.5 | 12 | 11.1 | 3 | 21.4 | 23 | 36.5 |
| Total | 288 | | 8 | | 108 | | 14 | | 63 | |

Table (2) presents results of stage 2 in terms of irregular inflection. As can be seen, we have (7) irregular inflections. To start with, there are 380 well-formed occurrences of inflections, and 200 ill-formed occurrences of inflections distributed as (7) instances of omission, 127 instances of addition, 36 instances of misselection and 30 instances of left undone.

**Table 2: Results collected from the participants at stage 2**

| | Irregular Inflection | | | | | | | | | |
|---|---|---|---|---|---|---|---|---|---|---|
| Items | Well-formed | | Ill-formed | | | | | | | |
| | | | Omission | | Addition | | Misselection | | Left undone | |
| | F | % | F | % | F | % | F | % | F | % |
| Consonant change | 46 | 12.1 | 4 | 57.1 | 12 | 9.4 | 4 | 11.1 | 3 | 10 |
| Vowel +Consonant change | 46 | 12.1 | 2 | 28.6 | 7 | 5.5 | 4 | 11.1 | 3 | 10 |
| Zero-morpheme | 72 | 18.9 | 0 | 0 | 39 | 30.7 | 7 | 19.4 | 6 | 20 |
| -A (plural) | 1 | 0.3 | 0 | 0 | 52 | 40.9 | 2 | 5.6 | 3 | 10 |
| Suppletion | 81 | 21.3 | 0 | 0 | 1 | 0.8 | 8 | 22.2 | 3 | 10 |
| -en (plural) | 36 | 9.5 | 0 | 0 | 1 | 0.8 | 0 | 0 | 0 | 0 |
| Vowel change | 98 | 25.8 | 1 | 14.3 | 15 | 11.8 | 11 | 30.6 | 12 | 40 |
| Total | 380 | | 7 | | 127 | | 36 | | 30 | |

Table (3) represents the results of the two stages in terms of irregular inflections as to make a comparison and to identify if there is a progress.

**Table 3: Comparison of the results at stage 1 and stage 2**

| | | Use of Irregular Inflections | | | | Chi-Square | P-Value |
|---|---|---|---|---|---|---|---|
| Inflections | | Stage one | | Stage two | | | |
| | | Well-formed | ill-formed | Well-formed | ill-formed | | |
| | F | 41 | 19 | 46 | 23 | 270.105[a] | .000 |

| | | | | | |
|---|---|---|---|---|---|
| Consonant change | % | 31.8% | 14.7% | 35.7% | 17.8% |
| Vowel +Consonant change | F | 37 | 23 | 46 | 16 |
| | % | 30.3% | 18.9% | 37.7% | 13.1% |
| Zero-morpheme | F | 69 | 51 | 72 | 52 |
| | % | 28.3% | 20.9% | 29.5% | 21.3% |
| -A (plural) | F | 5 | 55 | 1 | 57 |
| | % | 4.2% | 46.6% | 0.8% | 48.3% |
| Suppletion | F | 25 | 5 | 81 | 12 |
| | % | 20.3% | 4.1% | 65.9% | 9.8% |
| -en (plural) | F | 29 | 1 | 36 | 1 |
| | % | 43.3% | 1.5% | 53.7% | 1.5% |
| Vowel change | F | 82 | 39 | 98 | 39 |
| | % | 31.8% | 15.1% | 38.0% | 15.1% |
| Total | F | 288 | 193 | 380 | 200 |
| | % | 27.1% | 18.2% | 35.8% | 18.9% |

## 5.2. Discussion

This section is mainly concerned with discussing the results of the study and identifying the source of the errors made by Yemeni L1 Arabic learners of English either L1 transfer (Interlingual error), L2 learning strategies (Intralingual error), and unrecognized or unique error. The discussion is based on stage 1 data where L1 transfer is highly manifested than UG role does. Also, this section demonstrates the role of UG in resetting the L1 feature reassembly into the L2 counterparts. Stage 2 data shows that UG role is largely observed than L1 transfer.

### 5.2.1. L1 transfer

According to the FRH, adult L2 learners are able to master the knowledge of grammatical features, i.e., inflectional categories and their features, in two stages (cf. Lardiere, 2008, 2009): stage 1 is identified in transferring the L1 grammatical knowledge and features setting most notably when L2 learners are observed not to be able to produce well-formed expressions; rather, they produce ill-formed and erroneous features marking. As for stage 2, it will be discussed later in the study. Thus, the following sections are devoted to identifying source of errors in the L1 irregular inflections.

#### 5.2.1.1. Consonant change

As illustrated in table (1), the participants produced both well-formed and ill-formed instances of this inflectional morpheme. Their well-formed occurrences are double than the ill-formed ones, but the question is: where do these errors realized in ill-formed instances come from? And the answer to which can be stated below in (2).

2- **Wifes**

In (2), there is a reasonable error and the source of which is interlingual, i.e., Arabic phonological system. Here, the participants show their failure to produce the voiced labiodental fricative /v/ because this sound, simply, does not exist in Arabic; rather, they produce the Arabic voiceless labiodental fricative /f/ (Shormani & AlSohbani, 2015). Therefore, they transfer the voicelessness feature of the Arabic /f/, hence, pronouncing /v/" (p. 136).

### 5.2.1.2. Vowel and consonant change inflection

Vowel and consonant change inflection is another error the participants produced as table (1) shows and found a difficulty to realize it in its well-formed. Consider (3).

3- We **were leaved** before week.

In (3), the source of error is interlingual that could be located with the Arabic sentence in (4). The participants, in this example, transfers the Arabic structure, i.e. the auxiliary verb 'Kaana' and the main verb 'leave'. Here, the use of 'Kaana' is meant to indicate general past tense that happened once only.

4- kuna naḥnu χaaḍarna qabl ʔasbuuʕin

    were    we      leaved    before a week.

    'We left before a week'

### 5.2.1.3. Vowel change

As has been illustrated in table (1), vowel change scores the highest rank between group, i.e. irregular inflections, with an overall percentage of 28.5%, but the participants produced reasonable number of errors as in (5).

5- She **spoken** good English.

The source of error in (5) is interlingual in terms of form. Unlike English that uses the auxiliary verb along with the main verb to express the present perfect tense, Arabic never uses auxiliary verb as tense carrier; instead, it uses the particle "قد" along with the past tense to realize the perfective aspect. Therefore, the omission of this auxiliary comes from the Arabic structure.

### 5.2.2. L2 Influence (Intralingual Error)

In this section, the main concern is to identify the source of errors coming from the L2/target language, English and its type and nature. English, in our study, influences access to UG properties in acquiring English irregular inflections by Yemeni adult L2 learners. It is also set to examine why participants produce these variable or erroneous expressions.

### 5.2.2.1. Consonant change

Access to UG properties while acquiring the consonant change inflection is still delimited by intralingual sources, English, and the sentence in (6) is an instance of this delimitation/constraining.

6- He **spended** all his money.

In (6), the source of error is English, developmental, where the participants overapply the regular ed-past formation rule to the irregular verb 'spend'. Earlier in acquisition and a matter of fact, both L1 or L2 learners start mastering the rule-based inflections, and then, they employ the overgeneralization strategy as a learning strategy to inflect the irregular inflections.

**5.2.2.2. Vowel and consonant change**

Vowel and consonant change inflection is also a formidable learning task where the participants show different degrees of variability, and some instances of these are seen in (7&8).

7- I **leaved** my mother yesterday.
8- He **teached** me in school

In (7&8), the source of error is English where learners resort to using the overgeneralization strategy by applying the regular-based rule to add -ed to the end of regular verb to the irregular verbs as 'leave & teach' where vowel and consonant change is the rule to inflect such type of verbs.

**5.2.2.3. The zero-morpheme**

As table (3) illustrates, and though it scores the second highest rate of well-formed instances between the groups, the participants still observed to produce reasonable instances of ill-formed instances withing the group. Consider (9-12) to realize how English influentially affects access to UG properties relevant to this inflection.

9- **Deers**
10- **Informations**
11- **Hypothesises**
12- He **leted** her teach in the college.

In (9-12), the source is intralingual, English, and the reason for which is the lack of equivalent for such expressions in the L1 and no access to the morphological competence (Lardiere, 2008) pertinent to such inflection. Therefore, the learners resort to the overgeneralization strategy applying the regular rule-based plural inflection, adding **-s** and its variant forms to the irregular form, and the rule-based **-ed** past inflection to the zero-morpheme verb as 'let'. At this stage, the learners are still unaware of the morphological spell-out/ morphophonological realization feature of the verb 'let', therefore, they normally resort to the regular rules.

**5.2.2.4. The -a plural**

The -a plural is the least well-formed produced inflection the participants display in our study. Consider (13&14) to identify the source of such errors.

13- There are many **curriculums** to choose from.
14- **Phenomenons.**

The main source of the errors in (13&14) is intralingual, English, due to the rare exposure and use of such inflections. The only available and optional strategy the learners can resort to is the overgeneralization strategy, applying the regular rule-based plural inflection, adding -s to the irregular form on which their acquisition depends.

**5.2.2.5. The -en plural**

Though the participants manifest good rate in the production of well-formed instances of -en plural morpheme, they still produce persistent errors in the use of such inflection as in (15).

15- **Childs**.

The source of this error is surely intralingual, developmental where the learner uses the oversuppliance of the regular rule-based plural formation to the irregular one.

**5.2.2.6. Vowel change**

Vowel change inflection scores the highest rate of well-formed instances, but the participants still encounter difficulties in producing some other forms of this inflections as it is illustrated in (16&17).

16- Ali is one of the best **mans** in the office.
17- We **speaked** Spanish.

As alluded above, the source of such error comes from English where learners overgeneralize the regular rules of applying the -s plural and -ed past morphemes to irregular plural and past instance. These intralingual errors mark the hypothesis creating and testing stage (see Shormani, 2014, et seq) that signifies that learners are still internalizing the L2 system.

**5.2.3. UG and Feature Reassembling**

In sections 5.2.1 and 5.2.2., the study has been concerned with the role of L1 transfer, and L2 influence analyzing, discussing and exemplifying stage 1 data, and providing empirical evidence for such role in L2 acquisition. This section is mainly set to present and discuss empirical evidence ascertained from stage 2 data demonstrating the role of UG properties where the study's participants have been exposed to sufficient and efficient access to the linguistic input. Also, a one-way ANOVA analysis was utilized to identify differences between groups and within groups as presented in table (3). Most importantly, UG properties role is measured by comparing the number of well-formed instances with the ill-formed counterparts in both stages as well as the ability to produce well-formed instances in stage 2 that the participants failed to produce in stage 1.

As illustrated in table (3), the data shows that the UG role is enhanced in irregular inflection, and this is clearly supported by the high performance the participants show in terms of the high percentage of well-formed instances in stage 2 comparing to that in stage 1. Simplifying a bit, as illustrated in table (3) the frequency and percentage of irregular well-formed instances in stage 2 is (F 380, 35.8%) higher than it is in stage 1(F 288, 27.1%). As for the ill-formed instances, it is (F 200, 18.9%) in stage 2 and (F 193, 18.2%) in stage 1. It is obviously clear that learners' performance in stage 2 is better than in stage 1. Simply put, the data in table (3) indicates that there is a significant finding between groups, i.e., the well-formed and ill-formed instances in both stages. Also, the one-way ANOVA utilized reveals that there is a statistically significant difference between groups, stage 1 and stage 2, as $p < .006$; this strongly supports the conclusion that the participants are advancing in their progress, i.e., the acquisition of irregular inflections, and hence, they still have access to UG.

As illustrated in table (3) and calculating the difference between the well-formed and ill-formed instance, it is interestingly clear that the UG role is still present let alone the difference is not as it is expected. This presence can be observed with the participants' ability to avoid the oversuppliance of the regular inflections to produce the irregular counterparts let alone the influence played by the L1 transfer. As so far discussed, it is observed that the nature of errors the participants made in stage 1 can be attributed equally to the L2 influence and L1 transfer. These errors are assumed to be a matter of hypothesis creation and testing (see Shormani, 2014, et seq) where learners are internalizing the L2 morphological competence, and hence manifesting the role of UG properties. Thus, as alluded above, both L1transfer and L2 influence/developmental errors are observed to play a very substantial role in the initial stages of mastering English irregular inflections in the study at hand, let alone that this L1 transfer or L2 influence provides positive or negative evidence, but with sufficient and efficient exposure to the surrounding linguistic input, the UG role can be realized in learners' endeavor to reconfigure the already assembled L1 parameters onto the L2 counterpart or in the well-formed production of the errors made previously, stage 1 in our case. Concretizing this conclusion, we discuss some empirical evidence supporting the direct and full access to UG role manifested in stage 2 data exemplified in (18-22) displaying the acquisition of morphological competence pertinent to English irregular past feature in very complex constructions that they failed to manifest in stage 1.

18- What did Ali think that Samy fixed with a crowbar?
19- I visited Sana'a before two years. It was fantastic trip. I enjoyed all my time and visited my family. I sat there around 4 months, and I saw how much that's town is beautiful, and how it made me in good mood.
20- She taught me how to play the piano.
21- I spent a week in Spain.
22- They told us there is a really beautiful garden

What the set in (18-22) tells is that the participants have successfully acquired the morphological competence of the past inflection via manifesting long-distance agreement in structures characterized by many movements of Wh-phrase, realizing the Exceptional Case Marking verbs (ECM) like 'think' in (18) our case whose complement is an embedded clause, complement phrase (CP) whose tense feature is the past that goes hand in hand with the auxiliary "did", past tense carrier of the main clause's verb 'think'. Additionally, in (19), suppletion and vowel change, consonant change represented by the verbs 'was, sat, saw & made' are well realized respecting both tense sequence and shift between regular and irregular inflections.

In table (3), the role of UG is statistically significant in all irregular categories in stage 2, except in consonant change, zero-morpheme, and -a plural inflections where the progress is less than 5% for the consonant change, zero-inflections, and no progress at all observed for the *a*-plural . If the results illustrated and discussed so far affirm that L2 learners still have access to UG principles and parameters, then, the question is: why is it that the learners' performance is not the expected one? (Shormani & AlSohbani, 2015) whose answer is never to be the absence of UG role under any circumstances. The reasonable answer can be attributed to "lingering transfer effects" (Lardiere, 2008, p.14) on one hand, and several linguistic and non-linguistic factors constraining UG role on the other hand (Slabakova, 2016; Shormani, 2014, et seq; Shormani & AlSohbani, 2015).

Accessing UG can be constrained by linguistic aspects as the linguistic input L2 learners are provided with (Muñoz, 2014; Shormani, 2014, et seq; Shormani & AlSohbani, 2015; Slabakova, 2016), its presentation methods, "input modelizing" (Shormani, & AlSohbani, 2015), and the linguistic competence of teacher. Language growth is strikingly based on accessing high qualitative and quantitative linguistic input. Slabakova (2016) argues that comprehensible, and non-ambiguous input should be presented in a communicative situation. Input modelizing to learn irregular inflections should not be fully based on a connectionist model, or analogy model per se; rather, it should utilize a massive exposure model where learners are attended by abundant use of the irregular uses of inflections in context. Type of input, authentic/naturalist or classroom instruction, L2 learners are presented with seriously facilitates access to UG properties (see Shormani & AlSohbani, 2015; Shormani, 2014, et seq).

Practice is a very influential source of language acquisition, and its impact on language acquisition is linked to length of exposure and the teacher's competence. Here, the low performance of learners on consonant change, zero-morpheme, and -a plural inflections, in our study, can be attributed to limited allotted time of practice in the classroom (2-4 hours in a day) presented by incompetent teacher incapable to identify what is difficult to learn so that perfect practice is needed and what is universal and easy to learn so that no practice is needed as learners will master them for free (Slabakova, 2016). Also, according to the minimalist approach, the learners' low performance can be attributed to rare or infrequent exposure to a substantial and efficient amount of input "before these forms become entrenched and part of a child's repertoire, because they tend to rely on token frequency" (Blom et al. 2012; Paradis et al. 2011; Rispens & De Bree 2015; Blom, et al., 2021, p. 16).

To support our argument reached so far, Clark (2009) argues that both regular and irregular forms are subjected to one processing system instead of two processing systems, rule-based system for regular and association-based system for irregular. She supports this with the French verb and Arabic noun as evidence where the choice of regularity and irregularity "depends on tense, person, and conjugation in specific verbs" as it is the case with French verb and "the choice among many plural forms depends on certain features of the noun-type (see Plunkett & Nakisa 1997)" (p. 396) as it is the case with the Arabic noun. Put differently, if acquisition of irregular inflections is not UG-guided process, why do L2 learners start their learning by regularizing the irregular? In other words, why L2 learners make errors at the early stages and then, with sufficient and efficient exposure, they produce these irregular forms properly as it is the case with the participants with their well-formed instances in stage 2, in our case study. Thus, all these factors, linguistic, and non-linguistic, are serious barriers for Yemeni L2 learners of English, in our study, to fully and directly access UG properties, and hence, make it difficult to reassemble the L1 features into L2 configurations.

## 6. Conclusion and pedagogical implications

This study has investigated the acquisition of English irregular inflections by Yemeni L1 Arabic learners through the lens of the UG framework and FRH. The results clearly demonstrate a two-stage developmental pattern. In Stage 1, learners' performance was predominantly influenced by L1 transfer (interlingual errors) and intralingual developmental strategies such as overgeneralization. Errors in this stage primarily stemmed from negative transfer of Arabic phonological and morphological systems as well as the learners' reliance on L2 learning strategies. Examples include substituting /f/ for /v/ due to phonological gaps in Arabic and adding -ed or -s morphemes to irregular verbs and nouns, reflecting developmental tendencies.

In contrast, Stage 2 results show significant improvement, with a higher rate of well-formed irregular inflections and a statistically significant difference from Stage 1. This improvement supports the hypothesis that adult L2 learners maintain access to UG, which facilitates the reconfiguration of L1 grammatical features to align with L2 norms. The enhanced performance in Stage 2, particularly in producing complex constructions and correctly using morphological features that were previously problematic, indicates an internalization of L2 morphological competence (see also Slabakova, 2016).

In contrast, Stage 2 (Table 2) results show significant improvement, with a higher rate of well-formed irregular inflections and a statistically significant difference from Stage 1. This improvement supports the hypothesis that adult L2 learners maintain access to UG, which facilitates the reconfiguration of L1 grammatical features to align with L2 norms. The enhanced performance in Stage 2, particularly in producing complex constructions and correctly using morphological features that were previously problematic, indicates an internalization of L2 morphological competence (see also Slabakova, 2016).

As for the difficulties encountered by the participants of this study, findings reveal persistent difficulties in certain irregular inflection types, such as consonant change, zero-morpheme, and -a plural inflections, where learners showed limited improvement. These areas of weakness are attributed not to the absence of UG access but rather to L1 effects and external constraints on UG activation. These constraints include the quality and quantity of linguistic input, pedagogical practices, instructional time, and teacher competence. Effective and frequent exposure to authentic language input, as well as more informed and targeted teaching practices, are necessary to facilitate the full activation of UG principles and optimize L2 morphological acquisition. These findings confirm that adult L2 learners retain access to UG and can gradually reassemble L1 features into L2 equivalents, provided that they receive sufficient and appropriate linguistic input. Both L1 transfer and L2 developmental processes play crucial roles in the early stages of acquisition, while sustained and meaningful exposure leads to the emergence of UG-driven restructuring and more native-like use of irregular inflections.

Based on the findings and conclusions of this study, we strongly recommend the inclusion of contrastive morphology course from the very early stages whereby both regular and irregular inflections are focused on and presented with massive exposure to instances of the inflection's form and use in communicative situation as this seriously helps in identifying where difficulties might arise, and thus, more practice will be given to them (see also Alshawsh & Shormani, 2024). In addition, this course marks where L1 effect converges and where it diverges. Thus, where L1 transfer ends, UG role begins. Based on the bottleneck hypothesis (Slabakova, 2016), the study at hand strongly recommends the focus on language structure where L2 learners are invited to concentrate on "linguistic features if and when the classroom communicative activities and the negotiation of meaning demand these features" (p. 407). To put it the other way, teaching should be carried out in a clear and unambiguous context, and "intrinsically related to communication" (p.407), and thus, L2 learners could access UG and master the "morphological competence" our study's approach emphasizes on.

However, the study has some limitations. First, the study involves 30 students as participants. A more comprehensive study could involve all students in both levels to broaden its scope and purpose. Another limitation is that the study involves only Ibb University students. A broader study could also involve Yemeni University students from other universities to represent different strata of the Yemeni University students. A final limitation has to do with the levels involved, viz., the 3$^{rd}$ and 4$^{th}$ levels. A more representative study could involve 1-4 students to identify the access to UG in level 1, 2, 3, and 4, and how UG role develops in each level. We leave these issues for future research.